\documentclass{article}

\PassOptionsToPackage{numbers}{natbib}
\usepackage[preprint]{neurips_2024}

\usepackage[utf8]{inputenc} 
\usepackage[T1]{fontenc}    
\usepackage{hyperref}       
\usepackage{url}            
\usepackage{booktabs}       
\usepackage{amsfonts}       
\usepackage{nicefrac}       
\usepackage{microtype}      
\usepackage{xcolor}         
\usepackage{multirow}
 \usepackage{graphicx}
 \usepackage{amsmath} 
 \usepackage{placeins}
 \usepackage{float}
 \usepackage{animate}
\usepackage{siunitx}

\title{EchoPT: A Pretrained Transformer Architecture that Predicts 2D In-Air Sonar Images for Mobile Robotics}

\author{%
  Jan Steckel\thanks{First, Corresponding and Senior author on this paper}, \hspace{0.5ex}Wouter Jansen and Nico Huebel \\
  Cosys-lab, University of Antwerp, Antwerp, Belgium\\
  Flanders Make, Lommel, Belgium\\
  Email: \texttt{jan.steckel@uantwerpen.be}
}

\begin{document}

\maketitle

\begin{abstract}
The predictive brain hypothesis suggests that perception can be interpreted as the process of minimizing the error between predicted perception tokens generated by an internal world model and actual sensory input tokens. When implementing working examples of this hypothesis in the context of in-air sonar, significant difficulties arise due to the sparse nature of the reflection model that governs ultrasonic sensing. Despite these challenges, creating consistent world models using sonar data is crucial for implementing predictive processing of ultrasound data in robotics. In an effort to enable robust robot behavior using ultrasound as the sole exteroceptive sensor modality, this paper introduces EchoPT, a pretrained transformer architecture designed to predict 2D sonar images from previous sensory data and robot ego-motion information. We detail the transformer architecture that drives EchoPT and compare the performance of our model to several state-of-the-art techniques. In addition to presenting and evaluating our EchoPT model, we demonstrate the effectiveness of this predictive perception approach in two robotic tasks.
\end{abstract}

\section{Introduction}
Robots that navigate the real world often encounter noisy or incomplete sensor data, especially when using low-cost sensing modalities using long wavelengths, such as radar and sonar, and more so in industrial or agricultural environments. Indeed, industrial environments are littered with ultrasonic noise sources \cite{ullisch-nelkenAnalysisNoiseExposure2018}, such as compressed air \cite{schenckAirleakSlamDetectionPressurized2019}, CNC machining tools such as lathes and mills \cite{jozwikMonitoringNoiseEmitted2018}, and agricultural environments have high noise loads when handling grains, for example, \cite{zhangComparisonExperimentalMeasurements2020}, just to name a few. Despite the prevalence of these noise sources, one of the significant benefits of using ultrasound for robotics applications is the fact that it is virtually unhindered by medium distortions such as fog or dust particles \cite{steckel3dLocalizationBiomimetic2011}, as well as the low cost of relatively high-resolution 3D imaging sonar sensors \cite{moto:c:irua:165188_kers_a, moto:c:irua:168766_vere_urti,allevatoEmbeddedAircoupledUltrasonic2020, allevatoRealtime3DImaging2020}. These sensors can generate 2D or 3D range/direction representations of the environments, often referred to as energyscapes \cite{steckelBroadband3DSonar2012}, and have proven their application potential in mobile robotics. One interesting approach to mobile robotics using ultrasound is based on the subsumption architecture \cite{brooksRobustLayeredControl1986} and uses an analog to the widely known optical flow which called acoustic flow \cite{moto:c:irua:117369_pere_acou, steckelAcousticFlowBasedControl2017a,moto:c:irua:184471_jans_adap}. Acoustic flow is an approach to using expected transformations in the sensor observations, based on the robot's ego-motion data, to control a robot in a desired manner without explicit object segmentation.

When analyzing the acoustic flow equations given in \cite{moto:c:irua:117369_pere_acou, steckelAcousticFlowBasedControl2017a,moto:c:irua:184471_jans_adap}, one can interpret these as a special, closed form case of predictive processing. Predictive processing, or the predictive brain hypothesis, states that perception can be interpreted as the minimization of an error signal between a model-generated perception token and an observed token by altering the model's state \cite{clarkBustingOutPredictive2017, clarkWhateverNextPredictive2013}. As the acoustic flow model defines how sensor data changes and proposes closed-form solutions for testing predictions of sensor data evolution over time given a set of robot ego-motion parameters, acoustic flow can be seen as a form of predictive processing. While the acoustic flow model has proven to be a capable paradigm for robot navigation, issues remain with using acoustic flow as a prediction model. As a primary issue, the acoustic flow model ignores the concept of a point-spread function in the imaging sonar, ignoring aspects like the Rayleigh limit, which do govern real-world sonar sensors \cite{pailhasFullFieldView2017, paulEffectDSPPoint1993, steckelSparseDecompositionInair2014}. Furthermore, the acoustic flow model introduces deformations of the actual point-spread function when used for sensor data prediction, based on the non-isotropic forward prediction function \cite{steckelAcousticFlowBasedControl2017a}. 

To overcome these issues, we propose a learning-based model. This model is based on deep neural networks and transformers to perform sensor data prediction and is trained in a completely self-supervised manner, inspired by how large language models are trained \cite{achiamGpt4TechnicalReport2023, mikolovStrategiesTrainingLarge2011}. This self-supervision allows gathering vast amounts of sensor data without an expensive labeling step or the need for teacher sensor data. Indeed, the introduction of teacher sensor modalities has been done in previous works \cite{christensenBatvisionLearningSee2020a, gaoVisualechoesSpatialImage2020, paridaImageDepthImproving2021, schulteDeepLearnedAirCoupledUltrasonic2022}, in which either image enhancement was trained using a structured environment, or where a supervisory sensor modality is used (such as a camera or a 3D LiDAR) to produce a depth map to which the neural network regresses to using the sensor data. In our approach, called EchoPT (for the Echo-Predicting Pretrained Transformer), we use only the information from the sonar sensor, in conjunction with the desired robot ego-motion parameters set by the robot's controller, to predict future sensor frames, either in a single-shot or in an auto-regressive manner (where new predictions are made using previous predictions).

In the remainder of this paper, we will first detail the general architecture of the problem that EchoPT solves and give implementation details on our transformer-based neural network architecture. Next, we will provide exhaustive performance metrics on EchoPT and benchmark its prediction capabilities with several state-of-the-art methods. Next, we will illustrate the power of a system like EchoPT for predictive processing in the context of mobile robotics applications and the advantages that EchoPT has over existing methods. Finally, we will discuss the limitations of our work as well as the benefits that the EchoPT model provides and indicate areas of future work.

\section{EchoPT - the Echo-Predicting Pretrained Transformer}

\subsection{Problem Formulation and Experimental Setup}

\begin{figure*}
    \centering
    \includegraphics[width=\linewidth]{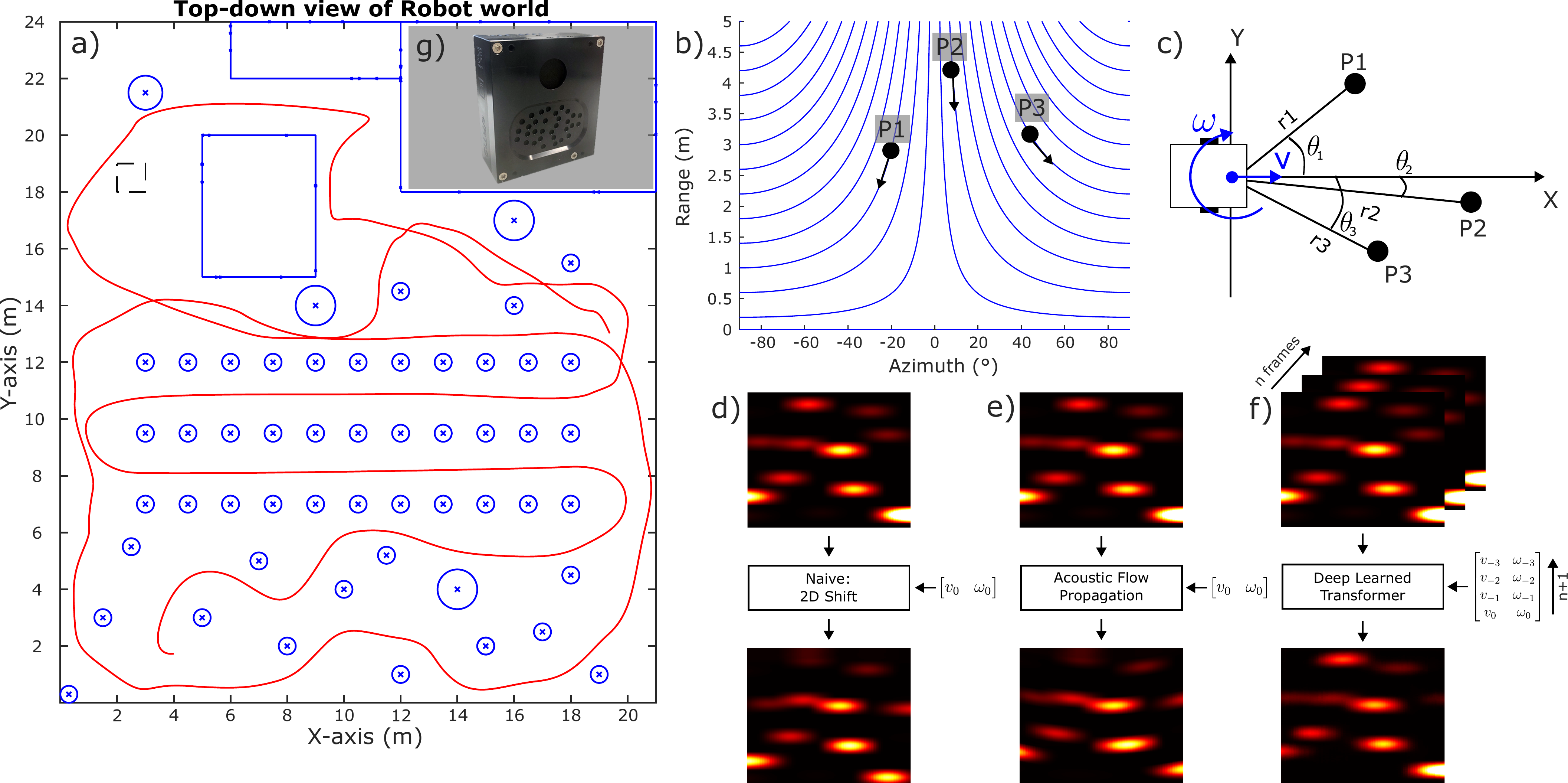}
    \caption{Overview of the experimental setup. Panel a) shows the simulation environment in which a two-wheeled robot drives. A sketch of the robot is shown in panel c). The robot uses an array-based imaging sonar sensor (panel g) capable of generating range-direction energy maps (called energyscapes), shown in panels d)-f). This sensor is modeled in the simulation environment based on accurate models of acoustic propagation and reflection. Panel b) shows what is called the acoustic flow model. This model predicts how objects in the sensor scene move through the perceptive field based on a certain robot motion. The blue flow lines are shown for a linear robot motion. Panels d)-f) show the task that is being solved in this paper: how can novel sensor views be synthesized given a certain set of robot velocity commands $\begin{bmatrix}v_{lin} & \omega_{r}\end{bmatrix}$. Panel d) shows the prediction based on the naive shifting of the image in the range and direction dimensions. Panel e) shows the operation using the acoustic flow model of panel b). Both these operators can only use the last frame to do the prediction. Panel f) shows the EchoPT model, which takes in $n$ previous frames and velocity commands and predicts the novel view using a transformer neural network. } 
    \label{fig:overview}
\end{figure*}

As a general concept, the EchoPT model aims to predict a novel sonar image token based on a sequence of previous sonar tokens and associated previous velocity commands executed by the robot and the velocity commands that the robot will perform in the next step. Figure \ref{fig:overview} shows an overview of this. The robot and a schematic view of the sonar sensor are depicted in panel c), showing the linear ($v_l$) and rotational ($\omega_r$) velocity components of the robot. Three reflectors are in the sensor's field of view, labeled P1, P2, and P3. Each of these is observed in a polar coordinate system consisting of range $r$ and azimuth direction $\theta$. Note that in this paper, we restrict ourselves to 2D sonar images in the horizontal plane, which has shown to be sufficient for many basic robotics tasks \cite{moto:c:irua:187594_jans_sona}. Extensions to 3D sonar sensors are currently not considered in this paper, as they require a non-uniform spatial sampling strategy based on the fact that sonar data lives on a non-euclidean manifold \cite{moto:c:irua:165339_reij_opti}. We will detail this limitation in section \ref{sec:limits}.

When the robot performs a particular set of velocity commands, the evolution of the sensor data over time can be calculated using the acoustic flow equations. More specifically, when performing a linear motion (without rotation components), the reflectors move over so-called flow lines, which are the solution of a set of differential equations (see \cite{moto:c:irua:117369_pere_acou}):
\begin{equation}
    r[t] \cdot sin( \theta[t] ) = R_c
\end{equation}
Here, $r[t]$ stands for the range on which the reflector is observed, and $\theta[t]$ is the azimuth direction. The constant $R_c$ is the range the flow-line intercepts with the $\theta=0$ direction. Exemplary flow lines can be seen in figure \ref{fig:overview}, panel b). 

To generate data for the experiments in this paper, we use the simulator developed in \cite{moto:c:irua:117369_pere_acou, steckelAcousticFlowBasedControl2017a,moto:c:irua:184471_jans_adap}, which has been validated to produce life-like sonar data for human-created environments, where robot algorithms have been trained in simulation and deployed using real-world sonar sensors without modifications to the processing pipeline. Using a simulation engine allows the generation of vast amounts of data and repeatable experiments on the effects of the implemented control algorithms, allowing statistical analysis of the results. The simulation environment in which the robot navigates is shown in figure \ref{fig:overview}, panel a). For benchmarking purposes, we implement three approaches to predict novel sensor tokens, all shown in panels d), e), and f). In panel d), we illustrate the most naive approach: for a given rotational and linear velocity, shift the images over the performed displacement along the according coordinate axes. This model is correct for rotational velocities but incorrect for linear velocity components, as seen from the acoustic flow equations in \cite{moto:c:irua:117369_pere_acou}. Furthermore, transmitter and receiver directivities are not taken into account in this approach. A more refined approach is illustrated in panel e), where we use the acoustic flow equations to predict the novel sensor views by transforming each image coordinate pixel to their new locations and then resampling the image on the original grid. This approach, however, still fails to incorporate spatial directivity patterns. Finally, panel f) shows the approach using EchoPT, which takes in a stack of $n$ previous sonar tokens and a set of previous velocity commands and predicts the next sonar token from that data and the planned velocity commands. Note that the naive and acoustic flow methods only process a single input image, as there is no precise formulation on using multiple input images to do that prediction with multiple input frames. Panel g) shows the sensor modeled in the simulation engine, an eRTIS imaging sonar sensor \cite{moto:c:irua:165188_kers_a} that generates the desired range/direction energy maps of the environment.

\subsection{The Architecture of EchoPT}
\begin{figure*}
    \centering
    \includegraphics[width=\linewidth]{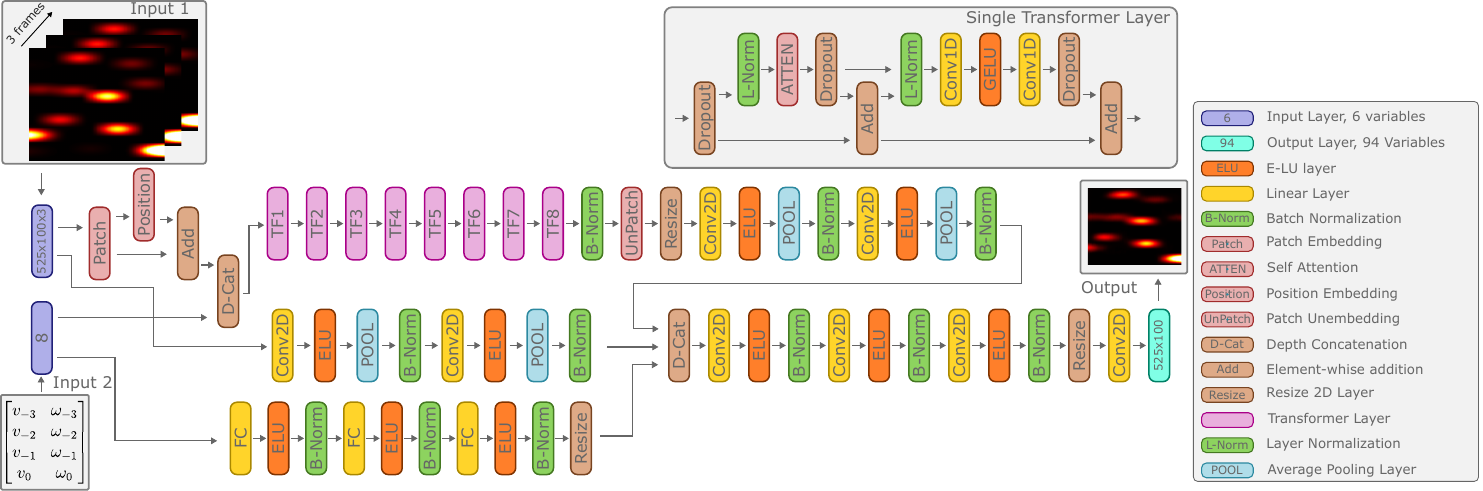}
    \caption{Overview of the network architecture of EchoPT. The EchoPT model has two inputs: the set of $n$ previous input frames (set to 3 in this paper) and the $n+1$ velocity commands (three previous and one for the prediction). The model has three main parallel branches: a transformer branch, a feed-forward convolutional branch for the sonar images, and an MLP pipeline using the velocity commands as input. These three branches are depth-concatenated and passed through more feed-forward convolutional layers to obtain a single output image.} 
    \label{fig:networkarch}
\end{figure*}

When designing the architecture of our EchoPT model, we were inspired by two mainstream model families. On the one hand, we took inspiration from large language models using transformers, which perform auto-regressive prediction of tokens (in our case, sonar images instead of text), trained in a self-supervised manner \cite{weiEmergentAbilitiesLarge2022}. On the other hand, we were inspired by vision transformers \cite{khanTransformersVisionSurvey2022, rajaniConvolutionalVisionTransformer2023, sunDPViTDualPathVision2022}, which use patch-embedding to embed the large input images into a lower-dimension latent space, which the transformer model subsequently processes. Similar approaches using the idea behind ViT have been used on 2D underwater sonar data for object recognition and detection \cite{raoVariousDegradationDual2024, sunDPViTDualPathVision2022a, yuRealtimeUnderwaterMaritime2021}, but not for the prediction of the following sensor token. The overall architecture of our model can be seen in figure \ref{fig:networkarch}. The model consists of three main branches: a branch with a transformer, using patch embedding and positional encoding to embed the input images into a latent space, which then gets concatenated with the velocity commands. This is passed through several transformer layers, each consisting of self-attention (6 heads and 384 key, query and value channels), and a non-linear transformation with a 500 dimensional latent space. After the transformer step, the patch embedding is reversed using the full transformer latent space, and the resulting 2D images are then passed through several feed-forward convolutional layers. In parallel, a feed-forward 2D convolutional pipeline operates on the stack of input images, and an MLP pipeline operates on the velocity commands. The outputs of these three branches are then resized to a uniform size and depth-concatenated. Finally, these concatenated representations are passed through a convolutional pipeline, yielding the final output image. The model, as implemented, has around 9 million learnable parameters, with the majority (7.8 million) situated in the eight transformer layers. The detailed architecture can be found in the source code posted on Zenodo \cite{echoPTZenodo} and in the appendix in table \ref{tab:learnable_params}. The model is implemented using the Matlab Deep Learning Toolbox version 2024a, using custom layers for the transformer implementation.

\subsection{Dataset and Training}
As stated before, we chose to build the dataset for this paper using the simulator that was developed in \cite{moto:c:irua:117369_pere_acou, steckelAcousticFlowBasedControl2017a,moto:c:irua:184471_jans_adap}. This simulator was validated with real-world measurements, as it has been used to develop and tune complex robotic control algorithms, which were then transferred to real-world robot experiments without changing the processing pipeline. Furthermore, using a simulator allows the generation of large datasets, as the whole data generation can happen in parallel and faster than in real-time. For this experiment, we generated ten datasets of 25 minutes. Each dataset contained 7500 sonar measurements measured at 5 Hz, with linear robot speeds between -0.3 m/s and 0.3 m/s and rotational speeds between -1 \si{\radian\per\second} and 1 \si{\radian\per\second}. In total, we simulated 5 hours of sonar data for the initial training and test dataset, consisting of 75000 sonar images. Before training, the dataset was pre-processed to build the input data stacks together with the velocity commands, and we used custom file data stores for easy access to the training data.

The model was trained using around 3.5 hours of sonar data (around 63000 sonar images) on a single NVIDIA RTX 4090 GPU and took around 36 hours. We used the Adam optimizer with a constant learning rate of 5e-5 for 1000 epochs with a mini-batch size of 64. The final network is picked by choosing the best validation loss after 1000 epochs, calculated on a validation set of 1600 image stacks (three input images, 8 velocity commands, and one output image) chosen randomly from the original dataset. Finally, all tests were done on an additional simulation run, which was not used during training. Details on the optimizer settings can be found in the appendix in table \ref{tab:optim}.

\section{Experimental Results}
In this section, we will detail several experimental results obtained using our EchoPT model. We will also detail several prediction experiments and benchmark them against the naive and acoustic flow method for next-token prediction. Furthermore, we will detail the power of predictive processing using EchoPT in two robotics experiments.

\subsection{Prediction Performance of EchoPT}
To evaluate the capabilities of EchoPT, we performed an experiment where the task of the model was to predict the next sonar sensor token in a sequence of images. Figure \ref{fig:prediction_corridor} in the appendix show an example of such a prediction. Figure \ref{fig:prediction_corridor_single} above shows a summery of this. In the figure in the appendix, the input sequence of images is seen in panels a)-c), indicated with T1, T2, and T3. The following image in the sequence is T4, which is the image that is to be predicted based on T1-T3. The prediction of EchoPT can be seen in panel e), labeled T4(Predicted). As the differences between subsequent frames are not that large, we calculated the differences between T4(Predicted) and the images T1-T4 and plotted them in panels f)-i). These difference images show that the difference between T4 and the predicted image is very low (panel i) and that there indeed is a difference between T4(Predicted) and T1-T3 due to the robot motion. We calculated the 2D correlation between T4(Predicted) and T1-T4, shown in panels j)-n), to further illustrate the prediction capabilities. Here, the blue cross indicates the middle of the correlogram, corresponding to no shift in the range or direction axes. The difference between the peak of the correlation function and the shift is largest when comparing the oldest images to the prediction and zero when comparing T4 to the prediction (panel m). This further illustrates the performance of the applied method.

\begin{figure*}
    \centering
    \includegraphics[width=\linewidth]{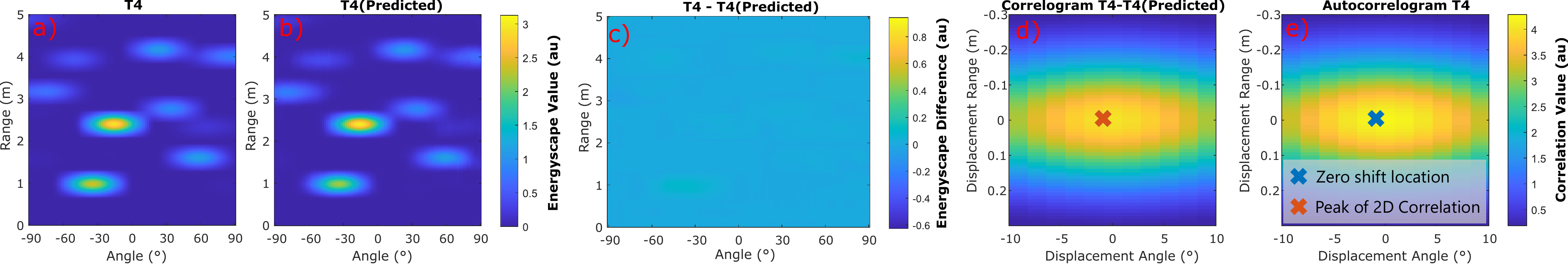}
    \caption{Condensed version of figure \ref{fig:prediction_corridor} in the appendix. Panel a) shows the target sonar image, and panel b) shows the predicted image. Panel c) shows the difference between the two images, and panels d) and e) show the 2D correlogram.} 
    \label{fig:prediction_corridor_single}
\end{figure*} 

\begin{figure*}
    \centering
    \includegraphics[width=\linewidth]{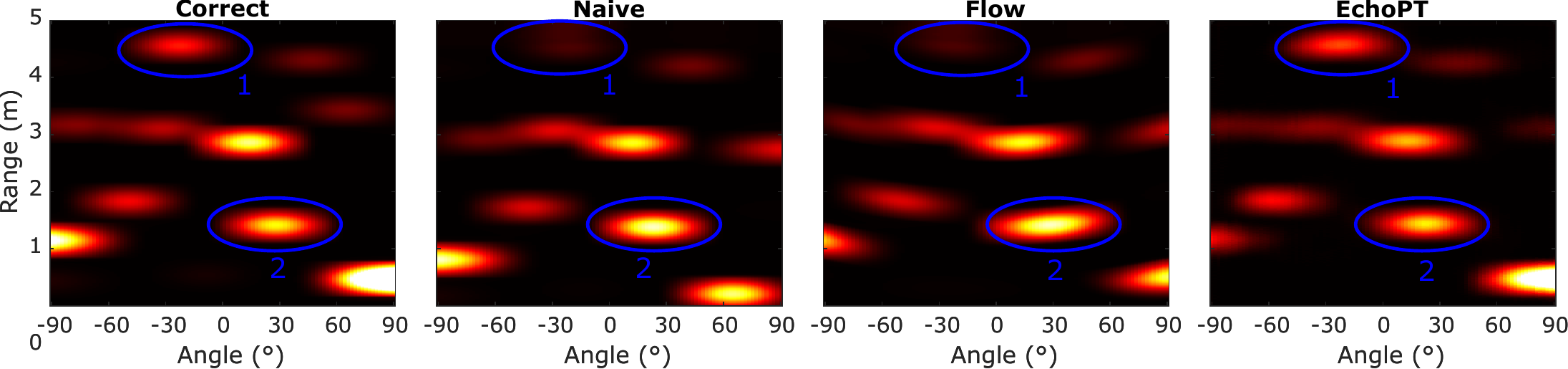}
    \caption{Prediction results of a single frame using three prediction methods: the naive operation, which shifts the image in the range and direction dimension; the Acoustic flow approach, which uses the acoustic flow equations to transform the image; and finally, the EchoPT prediction. } 
    \label{fig:ar_methods}
\end{figure*}

To benchmark the performance of EchoPT, we compared the predictive performance of EchoPT to two other methods, the naive method and the acoustic flow method, as explained in the previous section. Figure \ref{fig:ar_methods} shows an example of this comparison. We predicted the next sonar token using the three previously mentioned methods and plotted the resulting image. We highlighted two features (labeled 1 and 2) in the image, which show the main differences between the two approaches. The naive approach fails to incorporate intensity changes and local deformations (which is especially apparent in feature 1). In contrast, the acoustic flow approach deforms the imaging sensor's point-spread function (especially visible in feature 2). The point-spread function of an imaging sensor is an essential aspect of the performance of the sensor \cite{moto:c:irua:123137_stec_sona}, and altering this point-spread function can have detrimental effects in downstream tasks such as place-recognition \cite{moto:c:irua:133180_stec_spat}. In contrast, our EchoPT model successfully predicts the sensor data transformation while retaining all the major image features.

While the results in figure \ref{fig:ar_methods} are promising, they are illustrative and not evidence of a well-functioning prediction model. To further analyze the performance, we performed the token prediction using the three models on 1000 images from the test data set and calculated two performance metrics: the cross-correlation and the normalized root mean squared difference. We then calculated each test run's mean and standard deviation and collated this data in table \ref{combined-table}. In this table, the first column indicates the one-shot prediction error, which has been tested in this experiment, and we can conclude EchoPT model outperforms the other models in a statistically significant manner (2-sample $t$-test, p<0.001).

\subsection{Auto-regressive Prediction}
We performed an auto-regressive prediction task to test the model's capabilities further. In this task, the model is tasked to predict images several time steps ahead, using its previous predictions as input data. We tested three scenarios: AR-3, AR-5, and AR-10 on the same 1000 samples from the test set. The number behind the AR designator indicates how many frames in the future the model needs to predict (i.e., 3, 5 and 10). The one-shot approach would, in that line of reasoning, be AR-1. Exemplary results can be found in the appendix in figures \ref{fig:ar_methods_appendix} and \ref{fig:ar_diffs_appendix}, showing the predictions and the prediction differences, respectively. However, table \ref{combined-table} also shows the performance metrics of the three models for the three auto-regressive prediction tasks. Again, the EchoPT model outperforms the other models in all tasks in a statistically significant manner (2-sample $t$-test, p<0.001).

\begin{table}
  \caption{Performance Metrics for Sonar Prediction}
  \label{combined-table}
  \centering
  \begin{tabular}{lcccc}
    \toprule
    & \multicolumn{4}{c}{Cross Correlation [ $\mu (\sigma$) $\uparrow$ ]} \\
    \cmidrule(lr){2-5}
    Method & One-Shot & AR-3 & AR-5 & AR-10 \\
    \midrule
    Naive         & 0.85 (0.24) & 0.66 (0.31) & 0.56 (0.30) & 0.38 (0.29) \\
    Acoustic Flow & 0.86 (0.24) & 0.73 (0.33) & 0.69 (0.34) & 0.56 (0.33) \\
    \textbf{EchoPT (ours)}        & \textbf{0.97 (0.02)} & \textbf{0.83 (0.25)} & \textbf{0.80 (0.26)} & \textbf{0.69 (0.24)} \\
    \midrule
    & \multicolumn{4}{c}{Normalized Root Mean Squared Difference [ $\mu (\sigma$) $\downarrow$ ]} \\
    \cmidrule(lr){2-5}
    Method & One-Shot & AR-3 & AR-5 & AR-10 \\
    \midrule
    Naive         & 0.42 (0.37) & 0.76 (0.54) & 0.83 (0.44) & 1.34 (3.22) \\
    Acoustic Flow & 0.38 (0.37) & 0.62 (0.56) & 0.65 (0.61) & 0.99 (1.63) \\
    \textbf{EchoPT (ours)}        & \textbf{0.15 (0.11)} & \textbf{0.46 (0.45)} & \textbf{0.52 (0.31)} & \textbf{0.80 (1.07)} \\
    \bottomrule
  \end{tabular}
\end{table}

\subsection{Predictive Processing: Slip Detection}
\begin{figure*}
    \centering
    \includegraphics[width=\linewidth]{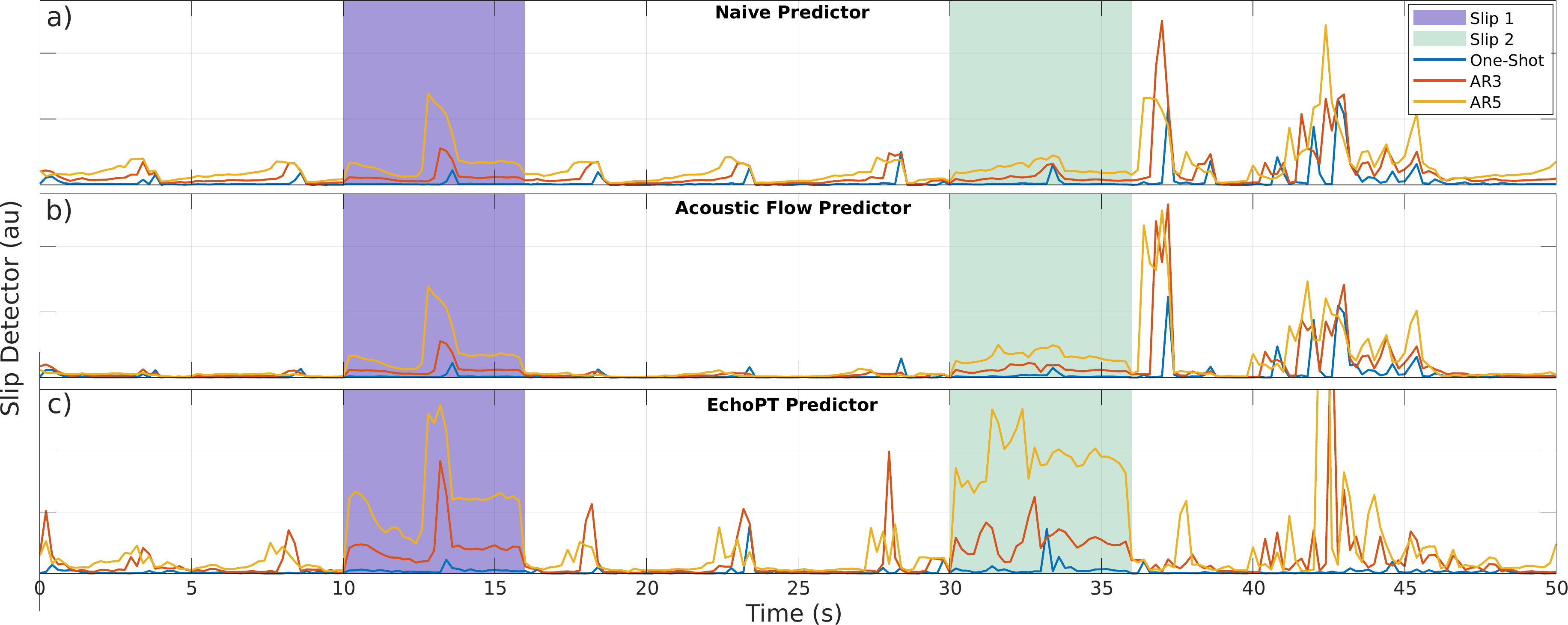}
    \caption{A first application of predictive processing in which a robot performs a trajectory in the environment from figure \ref{fig:overview}. In two periods (between 10s and 16s, and between 30s and 36s), the robot encounters slip conditions (meaning, the robot is not performing the motion that the robot expects to do). In the first section, the robot is slipping in both wheels; in the second condition, only one wheel slips. The plots show the slip detector, which uses differences in the predicted and measured sensor data for different prediction horizons (one shot, 3-frame auto-regressive, and 5-frame auto-regressive). Longer time horizons give the clearest slip detection signal, with EchoPT being the only one that detects the second slip condition. } 
    \label{fig:slipdetector}
\end{figure*}

As having a model that allows the prediction of sensor data tokens can be a powerful tool (as demonstrated expertly by large language models \cite{weiEmergentAbilitiesLarge2022}), we want to demonstrate the possible applications that this approach brings to robotics. In the first application, we use our EchoPT model to detect wheel slips in a mobile robot. Indeed, wheel slip is a challenging problem for outdoor robotics, where terrain conditions are not always predictable \cite{omuraWheelSlipClassification2017, ryuEvaluationCriterionWheeled2024}. To illustrate the usability of EchoPT in this application, we predict future sensor frames in an auto-regressive manner using sensor data, velocity commands, and previous predictions. We then calculate an error signal between the predicted frame and the measured frame and calculate an error signal $\epsilon(t)$ as follows:
\begin{equation}
    \epsilon(t) = \frac{\sqrt{\sum_{r,\theta}\bigg[ I_p(r,\theta) - I_m(r,\theta) \bigg]^2}}{\sqrt{|CC(I_p,I_m)|}}
\end{equation}
where $I_p$ is the predicted image, $I_m$ is the measured image, and $CC$ stands for the 2D correlation coefficient. The resulting error signals are calculated for three conditions, one-shot, AR-3, and AR-5, for the three prediction models (Naive, Acoustic Flow, and EchoPT). The results are shown in figure \ref{fig:slipdetector}, where each panel shows the results from the different predictors. We introduced two slip conditions, starting at 10s and 30s, wherein the first, both wheels of the robot slipped, and in the latter, only one wheel slipped. While all three detectors detect the first condition well, it is the second condition where EchoPT has the advantage, especially when using the 5-time step auto-regressive prediction (which, keep in mind, has no measured data in the final prediction step anymore and relies purely on previously predicted frames). The authors note that other techniques, such as using an IMU, might be a better solution to this particular problem, but the illustrative purpose of this application still stands.

\subsection{Predictive Processing: Robot Control}
\begin{figure*}
    \centering
    \includegraphics[width=\linewidth]{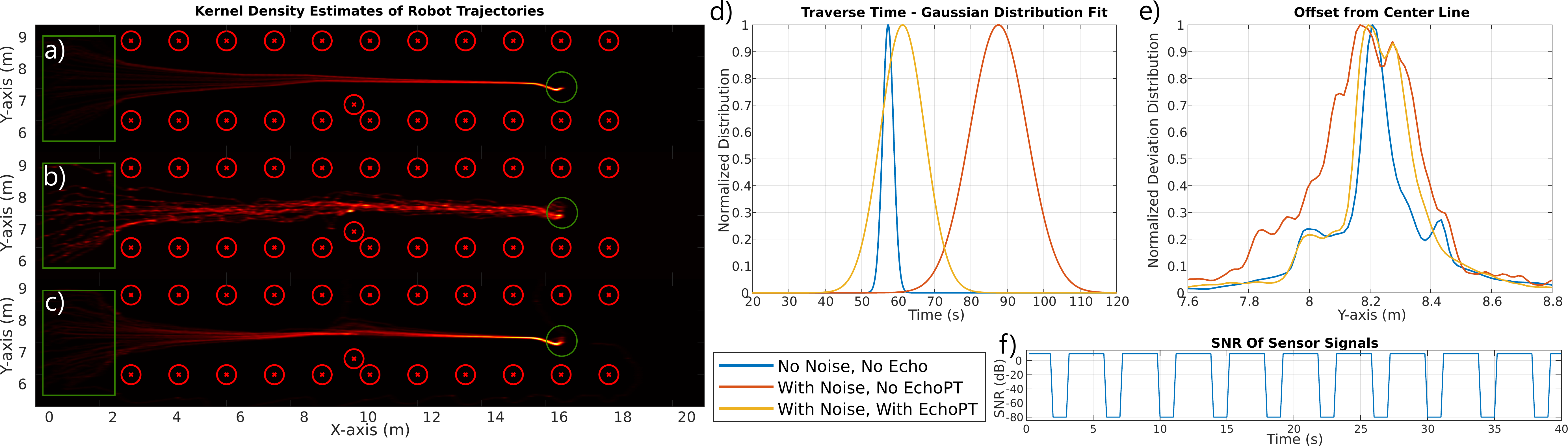}
    \caption{A second application of predictive processing in which a robot is tasked to drive from the green rectangular spawn boxes to the waypoint indicated by the green circles, using a subsumption-based control stack described in \cite{steckelAcousticFlowBasedControl2017a}. Panel a) shows the kernel density estimate of 50 runs with clean sensor data (SNR = 5dB). In panels b) and c), we added intermittent noise to the measured sensor data (shown in panel f, SNR = -80dB). In panel b), the original controller was used, showing the traversed paths' deterioration. In panel c), sensor data was predicted in an auto-regressive manner using EchoPT for the duration of the noise bursts and fed into the controller instead of the noisy data. Panel d) shows the travel time for the robot in the three conditions, showing a large increase in travel time for the controller from panel b. Panel e) shows the deviation from the midline of the corridor, again showing a large deviation when no predictive processing is used. Panel f) shows a small section of the evolution of the SNR over time.} 
    \label{fig:predproccontr}
\end{figure*}

A second, more challenging application is using our EchoPT model to overcome failures of sensor measurements due to high noise conditions. In industrial contexts, large bursts of ultrasonic noise might happen, completely overwhelming the faint echoes that the sensor picks up. Other failure modes might be intermittent loss of communication with the sensor, partial occlusions, etc. We set up an experiment where a robot drives through a corridor of circle reflectors, as shown in figure \ref{fig:predproccontr}. These reflectors, indicated by red circles, could represent trees in an orchard or the pillars of racks in an industry hall. The robot spawns with a random position and orientation in the green rectangular spawn boxes and drives toward the waypoint indicated by the green circle. The trajectory is repeated 50 times, and the robot locations are recorded. The robot uses the control algorithm detailed in \cite{moto:c:irua:117369_pere_acou, steckelAcousticFlowBasedControl2017a,moto:c:irua:184471_jans_adap}, which is based on the subsumption architecture and makes use of the acoustic flow model to perform corridor following. In panel a), we show the kernel density estimates of the robot trajectories for all 50 runs. The robot follows a stable trajectory through the corridor, further illustrated by the distribution of travel times (panel d) and the distribution of deviation from the middle of the corridor (panel e). 

Next, we introduce a sequence of high noise bursts, rendering the sensor data unintelligible during periods of 1.2 seconds (see panel f). The curve in panel f shows that approximately 30\% of the time the robot drives the data is lost due to the noise injection (changing the SNR from 5dB to -80dB). The resulting robot trajectory is shown in panel b, which shows the robot stopping due to the absence of sensible sensor data (indicated by the nodules in the kernel density estimates, as these correspond to the robot standing still). This is further illustrated by the significant increase in travel time (panel d) and the much larger deviation from the midline (panel e), indicating the robot controller has significant difficulties following a stable path through the corridor.

To overcome this problem, we apply the EchoPT model in auto-regressive mode to predict sensor data during the noise bursts based on previous measurements and the model's predictions. During the noise bursts, we feed the predicted sensor tokens from EchoPT into the robot controller, unaware of the noisy sensor conditions. The resulting robot paths are much more stable (panel c), approaching the performance of the noiseless case. This is also shown in panel d), with only a slight increase in transit time, and in panel e), with only a slight increase in deviation from the middle of the corridor.

\section{Limitations}
\label{sec:limits}
When contemplating the implementation of EchoPT and the experimental data illustrating its performance, it is not hard to come up with some limitations of our current presentation. For one, we are not using the concept of transformers in our system to its full potential. As the EchoPT model is working with a sequence of images, it would be wise to look for inspiration in how transformer architectures for video are constructed \cite{liSelftrainingMultisequenceLearning2022, ranasingheSelfsupervisedVideoTransformer2022, xieVideotrackLearningTrack2023}, by utilizing a two-tier transformer architecture, one for temporal, and one for spatial information. This is something we will be addressing in our future work, as this is a natural extension of the auto-regressive mode we have implemented now.

Another shortcoming is the use of 2D convolutions in the processing stack. When extending this idea of sensor data prediction to 3D, one inevitably stumbles into the fact that sonar data lives in a spherical manifold instead of a Cartesian one. Assuming a third dimension of the elevation direction and using a Cartesian grid to represent the data leads to severe over-representation of the data around the sphere's poles. Therefore, 2D convolutions are not a suitable operator to represent transformations in the 3D case. Solutions could be to replace the 2D convolutions with operators using spherical harmonics as basis functions \cite{poulenardEffectiveRotationinvariantPoint2019, varanasiDeepLearningFramework2020}, or to replace the convolutional blocks altogether with a transformer and use spatial patch embeddings which make use of patches distributed uniformly over the unit sphere with equal area \cite{choSamplingBasedSpherical2024, laiSphericalTransformerLidarbased2023, liuSphericalTransformerAdapting2021}.

In this study, we did not present any ablation experiments on the network architecture, which can be attributed to the limited availability of compute resources on the one hand, as well as the central message of this paper. Indeed, the main message is that transformer-like architectures can be used to perform predictive processing on in-air sonar data (which, to the best of the author's knowledge, is the first time this has been done). As we do not claim that the architecture of the EchoPT model is ideal in any way, the execution of ablation experiments was not part of the experiments in this paper. In terms of computing, a single training run currently takes around 36 hours, and a statistically meaningful ablation study would take around two weeks to compute, which is not available to the researchers at the time of writing. However, in short-term future work, these shortcomings will undoubtedly be addressed.

A final shortcoming of the current paper is the lack of real-world experiments, as all calculations were made using a simulator. This simulator was already used extensively in other papers \cite{moto:c:irua:117369_pere_acou, steckelAcousticFlowBasedControl2017a,moto:c:irua:184471_jans_adap}, including a validation of that simulator with real-world experiments. Therefore, we believe that using a simulator in this study is instead an advantage because it allows us to produce much more data and many more experiments, which then allows us to perform a detailed statistical analysis of the experimental data. Real-world experiments are also part of our future work.

\section{Conclusions}
In this paper, we presented our new EchoPT model, which stands for the Echo-Predicting Pretrained Transformer. We introduced the concept of predictive processing in robotics and the merits that this approach can bring to robotic perception in complex sensing scenarios. We thoroughly evaluated the performance of our EchoPT model in sensor token prediction tasks, both in single-shot and auto-regressive mode, and compared it to two other methods that solve the same prediction problem. The extensive tests we performed to quantify the performance of the models were analyzed using the appropriate statistical methods and indicate that our EchoPT model outperforms the benchmark models in a statistically significant manner.

In addition to solving sensor token predictions, we applied predictive processing in two robotic use cases: robot wheel slip detection and robotic motion control in high-noise environments. In the wheel slip detection, we built a detector signal that allows the detection of inconsistencies in sensory flow when the robot is not adhering to the underlying motion model (indicating wheel slip). Our EchoPT model outperforms the other two benchmark models in this task and is the only model that robustly detects the single-wheel slip condition. In the second task, a robot was challenged to navigate a corridor environment, where 30\% of the time, the robot's perception is rendered uninformative by adding vast quantities of noise (from 5dB SNR to -80dB SNR). Using auto-regressive prediction models based on our EchoPT prediction, the robot can still robustly navigate the corridor environment in which the robot is placed. The performance of this system was tested using 50 runs in each case.

We believe that with this paper, we have indicated the advantages deep-neural network architectures based on transformers can have for supporting predictive processing for neglected sensor modalities such as in-air sonar sensing. Indeed, the number of researchers working on vision or LiDAR-based systems for robotics is vastly greater than researchers working on in-air ultrasound for robotics (at the time of writing, searching for "robotics sonar ultrasound air" on Google Scholar since 2023 yields 425 papers while searching for "robotics vision LiDAR" yields 16.800 results). We hope that by writing this paper, we can inform the machine-learning community about this powerful modality, which is being used daily by millions of bats worldwide.

\clearpage
\section{Acknowledgements}
\setlength{\bibsep}{0.5ex}
\bibliographystyle{unsrtnat}
\bibliography{references_Zotero,additionalRefs}

\begin{thebibliography}{47}
\providecommand{\natexlab}[1]{#1}
\providecommand{\url}[1]{\texttt{#1}}
\expandafter\ifx\csname urlstyle\endcsname\relax
  \providecommand{\doi}[1]{doi: #1}\else
  \providecommand{\doi}{doi: \begingroup \urlstyle{rm}\Url}\fi

\bibitem[{Ullisch-Nelken} et~al.(2018){Ullisch-Nelken}, Kusserow, and Wol]{ullisch-nelkenAnalysisNoiseExposure2018}
Christian {Ullisch-Nelken}, Heiko Kusserow, and Andrea Wol.
\newblock Analysis of the noise exposure and the distribution of machine types at ultrasound related industrial workplaces in {{Germany}}.
\newblock \emph{Acta Acustica united with Acustica}, 104\penalty0 (5):\penalty0 733--736, 2018.

\bibitem[Schenck et~al.(2019)Schenck, Daems, and Steckel]{schenckAirleakSlamDetectionPressurized2019}
Anthony Schenck, Walter Daems, and Jan Steckel.
\newblock {{AirleakSlam}}: Detection of pressurized air leaks using passive ultrasonic sensors.
\newblock In \emph{2019 {{IEEE SENSORS}}}, pages 1--4. IEEE, 2019.
\newblock ISBN 1-72811-634-1.

\bibitem[J{\'o}zwik et~al.(2018)J{\'o}zwik, {Wac-W{\l}odarczyk}, Micha{\l}owska, and K{\l}oczko]{jozwikMonitoringNoiseEmitted2018}
Jerzy J{\'o}zwik, Andrzej {Wac-W{\l}odarczyk}, Joanna Micha{\l}owska, and Monika K{\l}oczko.
\newblock Monitoring of the noise emitted by machine tools in industrial conditions.
\newblock \emph{Journal of Ecological Engineering}, 19\penalty0 (1):\penalty0 83--93, 2018.

\bibitem[Zhang et~al.(2020)Zhang, Song, Li, and Zhong]{zhangComparisonExperimentalMeasurements2020}
Jie Zhang, Yongfeng Song, Xiongbing Li, and ChengHuan Zhong.
\newblock Comparison of experimental measurements of material grain size using ultrasound.
\newblock \emph{Journal of Nondestructive Evaluation}, 39\penalty0 (2):\penalty0 30, 2020.

\bibitem[Steckel et~al.(2011)Steckel, Vanduren, and Peremans]{steckel3dLocalizationBiomimetic2011}
Jan Steckel, Wouter Vanduren, and Herbert Peremans.
\newblock 3d localization by a biomimetic sonar system in a fire-fighting application.
\newblock In \emph{2011 4th {{International Congress}} on {{Image}} and {{Signal Processing}}}, volume~5, pages 2549--2553. IEEE, 2011.
\newblock ISBN 1-4244-9306-4.

\bibitem[Kerstens et~al.(2019)Kerstens, Laurijssen, and Steckel]{moto:c:irua:165188_kers_a}
Robin Kerstens, Dennis Laurijssen, and Jan Steckel.
\newblock {{eRTIS}}: A fully embedded real time {{3D}} imaging sonar sensor for robotic applications.
\newblock In \emph{{{IEEE}} International Conference on Robotics and Automation}, pages 1438--1443. Ieee, 2019.
\newblock ISBN 978-1-5386-6026-3.
\newblock \doi{10.1109/ICRA.2019.8794419}.

\bibitem[Verellen et~al.(2020)Verellen, Kerstens, Laurijssen, and Steckel]{moto:c:irua:168766_vere_urti}
Thomas Verellen, Robin Kerstens, Dennis Laurijssen, and Jan Steckel.
\newblock {{URTIS}} : A small {{3D}} imaging sonar sensor for robotic applications.
\newblock In \emph{Proceedings of the ... {{IEEE International Conference}} on {{Acoustics}}, {{Speech}}, and {{Signal Processing}}}, pages 4801--4805. IEEE, 2020.
\newblock ISBN 978-1-5090-6631-5.
\newblock \doi{10.1109/ICASSP40776.2020.9053536}.

\bibitem[Allevato et~al.(2020{\natexlab{a}})Allevato, Rutsch, Hinrichs, Pesavento, and Kupnik]{allevatoEmbeddedAircoupledUltrasonic2020}
Gianni Allevato, Matthias Rutsch, Jan Hinrichs, Marius Pesavento, and Mario Kupnik.
\newblock Embedded air-coupled ultrasonic {{3D}} sonar system with {{GPU}} acceleration.
\newblock In \emph{2020 {{IEEE SENSORS}}}, pages 1--4. IEEE, 2020{\natexlab{a}}.
\newblock ISBN 1-72816-801-5.

\bibitem[Allevato et~al.(2020{\natexlab{b}})Allevato, Hinrichs, Rutsch, Adler, J{\"a}ger, Pesavento, and Kupnik]{allevatoRealtime3DImaging2020}
Gianni Allevato, Jan Hinrichs, Matthias Rutsch, Jan~Philipp Adler, Axel J{\"a}ger, Marius Pesavento, and Mario Kupnik.
\newblock Real-time 3-{{D}} imaging using an air-coupled ultrasonic phased-array.
\newblock \emph{IEEE transactions on ultrasonics, ferroelectrics, and frequency control}, 68\penalty0 (3):\penalty0 796--806, 2020{\natexlab{b}}.

\bibitem[Steckel et~al.(2012)Steckel, Boen, and Peremans]{steckelBroadband3DSonar2012}
Jan Steckel, Andre Boen, and Herbert Peremans.
\newblock Broadband 3-{{D}} sonar system using a sparse array for indoor navigation.
\newblock \emph{IEEE Transactions on Robotics}, 29\penalty0 (1):\penalty0 161--171, 2012.
\newblock ISSN 1552-3098.

\bibitem[Brooks(1986)]{brooksRobustLayeredControl1986}
Rodney Brooks.
\newblock A robust layered control system for a mobile robot.
\newblock \emph{IEEE journal on robotics and automation}, 2\penalty0 (1):\penalty0 14--23, 1986.

\bibitem[Peremans and Steckel(2014)]{moto:c:irua:117369_pere_acou}
Herbert Peremans and Jan Steckel.
\newblock Acoustic flow for robot motion control.
\newblock In \emph{{{IEEE}} International Conference on Robotics and Automation}, pages 316--321. 2014.

\bibitem[Steckel and Peremans(2017)]{steckelAcousticFlowBasedControl2017a}
Jan Steckel and Herbert Peremans.
\newblock Acoustic {{Flow-Based Control}} of a {{Mobile Platform Using}} a {{3D Sonar Sensor}}.
\newblock \emph{IEEE Sensors Journal}, 17\penalty0 (10):\penalty0 3131--3141, May 2017.
\newblock ISSN 1530437X.
\newblock \doi{10.1109/JSEN.2017.2688476}.

\bibitem[Jansen et~al.(2021)Jansen, Laurijssen, and Steckel]{moto:c:irua:184471_jans_adap}
Wouter Jansen, Dennis Laurijssen, and Jan Steckel.
\newblock Adaptive acoustic flow-based navigation with {{3D}} sonar sensor fusion.
\newblock In \emph{Indoor {{Positioning}} and {{Indoor Navigation}} ({{IPIN}}), {{International Conference}} On}, pages 1--8. 2021.
\newblock ISBN 978-1-66540-402-0.
\newblock \doi{10.1109/IPIN51156.2021.9662596}.

\bibitem[Clark(2017)]{clarkBustingOutPredictive2017}
Andy Clark.
\newblock Busting out: {{Predictive}} brains, embodied minds, and the puzzle of the evidentiary veil.
\newblock \emph{No{\^u}s}, 51\penalty0 (4):\penalty0 727--753, 2017.

\bibitem[Clark(2013)]{clarkWhateverNextPredictive2013}
Andy Clark.
\newblock Whatever next? {{Predictive}} brains, situated agents, and the future of cognitive science.
\newblock \emph{Behavioral and brain sciences}, 36\penalty0 (3):\penalty0 181--204, 2013.

\bibitem[Pailhas et~al.(2017)Pailhas, Petillot, and Mulgrew]{pailhasFullFieldView2017}
Yan Pailhas, Yvan Petillot, and Bernard Mulgrew.
\newblock Full field of view point spread function for circular synthetic aperture sonar systems.
\newblock In \emph{Proceedings of {{Meetings}} on {{Acoustics}}}, volume~30. AIP Publishing, 2017.

\bibitem[Paul et~al.(1993)Paul, McHugh, and Shaw]{paulEffectDSPPoint1993}
J.~G. Paul, R.~McHugh, and S.~Shaw.
\newblock The effect of {{DSP}} on the point spread function of a sonar beamformer.
\newblock In \emph{International {{Conference}} on {{Acoustic Sensing}} and {{Imaging}}, 1993.}, pages 23--27. IET, 1993.
\newblock ISBN 0-85296-575-3.

\bibitem[Steckel and Peremans(2014)]{steckelSparseDecompositionInair2014}
Jan Steckel and Herbert Peremans.
\newblock Sparse decomposition of in-air sonar images for object localization.
\newblock In \emph{{{SENSORS}}, 2014 {{IEEE}}}, pages 1356--1359. IEEE, 2014.
\newblock ISBN 1-4799-0162-8.

\bibitem[Achiam et~al.(2023)Achiam, Adler, Agarwal, Ahmad, Akkaya, Aleman, Almeida, Altenschmidt, Altman, and Anadkat]{achiamGpt4TechnicalReport2023}
Josh Achiam, Steven Adler, Sandhini Agarwal, Lama Ahmad, Ilge Akkaya, Florencia~Leoni Aleman, Diogo Almeida, Janko Altenschmidt, Sam Altman, and Shyamal Anadkat.
\newblock Gpt-4 technical report.
\newblock \emph{arXiv preprint arXiv:2303.08774}, 2023.

\bibitem[Mikolov et~al.(2011)Mikolov, Deoras, Povey, Burget, and {\v C}ernock{\'y}]{mikolovStrategiesTrainingLarge2011}
Tom{\'a}{\v s} Mikolov, Anoop Deoras, Daniel Povey, Luk{\'a}{\v s} Burget, and Jan {\v C}ernock{\'y}.
\newblock Strategies for training large scale neural network language models.
\newblock In \emph{2011 {{IEEE Workshop}} on {{Automatic Speech Recognition}} \& {{Understanding}}}, pages 196--201. IEEE, 2011.
\newblock ISBN 1-4673-0367-4.

\bibitem[Christensen et~al.(2020)Christensen, Hornauer, and Stella]{christensenBatvisionLearningSee2020a}
Jesper~Haahr Christensen, Sascha Hornauer, and X.~Yu Stella.
\newblock Batvision: {{Learning}} to see 3d spatial layout with two ears.
\newblock In \emph{2020 {{IEEE International Conference}} on {{Robotics}} and {{Automation}} ({{ICRA}})}, pages 1581--1587. IEEE, 2020.
\newblock ISBN 1-72817-395-7.

\bibitem[Gao et~al.(2020)Gao, Chen, {Al-Halah}, Schissler, and Grauman]{gaoVisualechoesSpatialImage2020}
Ruohan Gao, Changan Chen, Ziad {Al-Halah}, Carl Schissler, and Kristen Grauman.
\newblock Visualechoes: {{Spatial}} image representation learning through echolocation.
\newblock In \emph{Computer {{Vision}}--{{ECCV}} 2020: 16th {{European Conference}}, {{Glasgow}}, {{UK}}, {{August}} 23--28, 2020, {{Proceedings}}, {{Part IX}} 16}, pages 658--676. Springer, 2020.
\newblock ISBN 3-030-58544-1.

\bibitem[Parida et~al.(2021)Parida, Srivastava, and Sharma]{paridaImageDepthImproving2021}
Kranti~Kumar Parida, Siddharth Srivastava, and Gaurav Sharma.
\newblock Beyond image to depth: {{Improving}} depth prediction using echoes.
\newblock In \emph{Proceedings of the {{IEEE}}/{{CVF Conference}} on {{Computer Vision}} and {{Pattern Recognition}}}, pages 8268--8277, 2021.

\bibitem[Schulte et~al.(2022)Schulte, Allevato, Haugwitz, and Kupnik]{schulteDeepLearnedAirCoupledUltrasonic2022}
Stefan Schulte, Gianni Allevato, Christoph Haugwitz, and Mario Kupnik.
\newblock Deep-{{Learned Air-Coupled Ultrasonic Sonar Image Enhancement}} and {{Object Localization}}.
\newblock In \emph{2022 {{IEEE Sensors}}}, pages 01--04. IEEE, 2022.
\newblock ISBN 1-66548-464-0.

\bibitem[Jansen et~al.(2022)Jansen, Laurijssen, and Steckel]{moto:c:irua:187594_jans_sona}
Wouter Jansen, Dennis Laurijssen, and Jan Steckel.
\newblock Real-time sonar fusion for layered navigation controller.
\newblock \emph{Sensors}, 22\penalty0 (9):\penalty0 1--18, 2022.
\newblock ISSN 1424-8220.
\newblock \doi{10.3390/S22093109}.

\bibitem[Reijniers et~al.(2020)Reijniers, Kerstens, and Steckel]{moto:c:irua:165339_reij_opti}
Jonas Reijniers, Robin Kerstens, and Jan Steckel.
\newblock An optimized spatial sampling strategy for wide-view planar array {{3D}} sonar sensors.
\newblock \emph{IEEE transactions on ultrasonics, ferroelectrics and frequency control}, 67\penalty0 (6):\penalty0 1236--1241, 2020.
\newblock ISSN 0885-3010.
\newblock \doi{10.1109/TUFFC.2020.2964991}.

\bibitem[Wei et~al.(2022)Wei, Tay, Bommasani, Raffel, Zoph, Borgeaud, Yogatama, Bosma, Zhou, and Metzler]{weiEmergentAbilitiesLarge2022}
Jason Wei, Yi~Tay, Rishi Bommasani, Colin Raffel, Barret Zoph, Sebastian Borgeaud, Dani Yogatama, Maarten Bosma, Denny Zhou, and Donald Metzler.
\newblock Emergent abilities of large language models.
\newblock \emph{arXiv preprint arXiv:2206.07682}, 2022.

\bibitem[Khan et~al.(2022)Khan, Naseer, Hayat, Zamir, Khan, and Shah]{khanTransformersVisionSurvey2022}
Salman Khan, Muzammal Naseer, Munawar Hayat, Syed~Waqas Zamir, Fahad~Shahbaz Khan, and Mubarak Shah.
\newblock Transformers in vision: {{A}} survey.
\newblock \emph{ACM computing surveys (CSUR)}, 54\penalty0 (10s):\penalty0 1--41, 2022.

\bibitem[Rajani et~al.(2023)Rajani, Gracias, and Garcia]{rajaniConvolutionalVisionTransformer2023}
Hayat Rajani, Nuno Gracias, and Rafael Garcia.
\newblock A convolutional vision transformer for semantic segmentation of side-scan sonar data.
\newblock \emph{Ocean Engineering}, 286:\penalty0 115647, 2023.

\bibitem[Sun et~al.(2022{\natexlab{a}})Sun, Zheng, Zhang, Ren, Xu, and Xu]{sunDPViTDualPathVision2022}
Yushan Sun, Haotian Zheng, Guocheng Zhang, Jingfei Ren, Hao Xu, and Chao Xu.
\newblock {{DP-ViT}}: {{A Dual-Path Vision Transformer}} for {{Real-Time Sonar Target Detection}}.
\newblock \emph{Remote Sensing}, 14\penalty0 (22):\penalty0 5807, 2022{\natexlab{a}}.

\bibitem[Rao et~al.(2024)Rao, Peng, Chen, and Tian]{raoVariousDegradationDual2024}
Jiahao Rao, Yini Peng, Jun Chen, and Xin Tian.
\newblock Various {{Degradation}}: {{Dual Cross-Refinement Transformer For Blind Sonar Image Super-Resolution}}.
\newblock \emph{IEEE Transactions on Geoscience and Remote Sensing}, 2024.

\bibitem[Sun et~al.(2022{\natexlab{b}})Sun, Zheng, Zhang, Ren, Xu, and Xu]{sunDPViTDualPathVision2022a}
Yushan Sun, Haotian Zheng, Guocheng Zhang, Jingfei Ren, Hao Xu, and Chao Xu.
\newblock {{DP-ViT}}: {{A Dual-Path Vision Transformer}} for {{Real-Time Sonar Target Detection}}.
\newblock \emph{Remote Sensing}, 14\penalty0 (22):\penalty0 5807, 2022{\natexlab{b}}.

\bibitem[Yu et~al.(2021)Yu, Zhao, Gong, Huang, Zheng, and Ma]{yuRealtimeUnderwaterMaritime2021}
Yongcan Yu, Jianhu Zhao, Quanhua Gong, Chao Huang, Gen Zheng, and Jinye Ma.
\newblock Real-time underwater maritime object detection in side-scan sonar images based on transformer-{{YOLOv5}}.
\newblock \emph{Remote Sensing}, 13\penalty0 (18):\penalty0 3555, 2021.

\bibitem[Anonymous(2024)]{echoPTZenodo}
Anonymous Anonymous.
\newblock Echopt: A pretrained transformer architecture for predicting 2d in-air sonar images in mobile robotics, May 2024.
\newblock URL \url{https://zenodo.org/record/11191954}.

\bibitem[Steckel(2015)]{moto:c:irua:123137_stec_sona}
Jan Steckel.
\newblock Sonar system combining an emitter array with a sparse receiver array for air-coupled applications.
\newblock \emph{IEEE sensors journal}, 15\penalty0 (6):\penalty0 3446--3452, 2015.
\newblock ISSN 1530-437X.
\newblock \doi{10.1109/JSEN.2015.2391290}.

\bibitem[Steckel and Peremans(2015)]{moto:c:irua:133180_stec_spat}
Jan Steckel and Herbert Peremans.
\newblock Spatial sampling strategy for a {{3D}} sonar sensor supporting {{BatSLAM}}.
\newblock In \emph{Proceedings of the {{International Conference}} on {{Intelligent Robots}} and {{Systems}}}, pages 723--728. Ieee, 2015.
\newblock ISBN 978-1-4799-9994-1.
\newblock \doi{10.1109/IROS.2015.7353452}.

\bibitem[Omura and Ishigami(2017)]{omuraWheelSlipClassification2017}
Takuya Omura and Genya Ishigami.
\newblock Wheel slip classification method for mobile robot in sandy terrain using in-wheel sensor.
\newblock \emph{Journal of Robotics and Mechatronics}, 29\penalty0 (5):\penalty0 902--910, 2017.

\bibitem[Ryu et~al.(2024)Ryu, Won, Chae, Kim, and Seo]{ryuEvaluationCriterionWheeled2024}
Sijun Ryu, Jeeho Won, Hobyeung Chae, Hwa~Soo Kim, and TaeWon Seo.
\newblock Evaluation {{Criterion}} of {{Wheeled Mobile Robotic Platforms}} on {{Grounds}}: {{A Survey}}.
\newblock \emph{International Journal of Precision Engineering and Manufacturing}, 25\penalty0 (3):\penalty0 675--686, 2024.

\bibitem[Li et~al.(2022)Li, Liu, and Jiao]{liSelftrainingMultisequenceLearning2022}
Shuo Li, Fang Liu, and Licheng Jiao.
\newblock Self-training multi-sequence learning with transformer for weakly supervised video anomaly detection.
\newblock In \emph{Proceedings of the {{AAAI Conference}} on {{Artificial Intelligence}}}, volume~36, pages 1395--1403, 2022.
\newblock ISBN 2374-3468.

\bibitem[Ranasinghe et~al.(2022)Ranasinghe, Naseer, Khan, Khan, and Ryoo]{ranasingheSelfsupervisedVideoTransformer2022}
Kanchana Ranasinghe, Muzammal Naseer, Salman Khan, Fahad~Shahbaz Khan, and Michael~S. Ryoo.
\newblock Self-supervised video transformer.
\newblock In \emph{Proceedings of the {{IEEE}}/{{CVF Conference}} on {{Computer Vision}} and {{Pattern Recognition}}}, pages 2874--2884, 2022.

\bibitem[Xie et~al.(2023)Xie, Chu, Li, Lu, and Ma]{xieVideotrackLearningTrack2023}
Fei Xie, Lei Chu, Jiahao Li, Yan Lu, and Chao Ma.
\newblock Videotrack: {{Learning}} to track objects via video transformer.
\newblock In \emph{Proceedings of the {{IEEE}}/{{CVF Conference}} on {{Computer Vision}} and {{Pattern Recognition}}}, pages 22826--22835, 2023.

\bibitem[Poulenard et~al.(2019)Poulenard, Rakotosaona, Ponty, and Ovsjanikov]{poulenardEffectiveRotationinvariantPoint2019}
Adrien Poulenard, Marie-Julie Rakotosaona, Yann Ponty, and Maks Ovsjanikov.
\newblock Effective rotation-invariant point cnn with spherical harmonics kernels.
\newblock In \emph{2019 {{International Conference}} on {{3D Vision}} ({{3DV}})}, pages 47--56. IEEE, 2019.
\newblock ISBN 1-72813-131-6.

\bibitem[Varanasi et~al.(2020)Varanasi, Gupta, and Hegde]{varanasiDeepLearningFramework2020}
Vishnuvardhan Varanasi, Harshit Gupta, and Rajesh~M. Hegde.
\newblock A deep learning framework for robust {{DOA}} estimation using spherical harmonic decomposition.
\newblock \emph{IEEE/ACM Transactions on Audio, Speech, and Language Processing}, 28:\penalty0 1248--1259, 2020.

\bibitem[Cho et~al.(2024)Cho, Jung, and Kwon]{choSamplingBasedSpherical2024}
Sungmin Cho, Raehyuk Jung, and Junseok Kwon.
\newblock Sampling based spherical transformer for 360 degree image classification.
\newblock \emph{Expert Systems with Applications}, 238:\penalty0 121853, 2024.

\bibitem[Lai et~al.(2023)Lai, Chen, Lu, Liu, and Jia]{laiSphericalTransformerLidarbased2023}
Xin Lai, Yukang Chen, Fanbin Lu, Jianhui Liu, and Jiaya Jia.
\newblock Spherical transformer for lidar-based 3d recognition.
\newblock In \emph{Proceedings of the {{IEEE}}/{{CVF Conference}} on {{Computer Vision}} and {{Pattern Recognition}}}, pages 17545--17555, 2023.

\bibitem[Liu et~al.(2021)Liu, Wang, Du, and Cai]{liuSphericalTransformerAdapting2021}
Yuqi Liu, Yin Wang, Haikuan Du, and Shen Cai.
\newblock Spherical transformer: {{Adapting}} spherical signal to {{CNNs}}.
\newblock \emph{arXiv preprint arXiv:2101.03848}, 2021.

\end{thebibliography}

\clearpage
\appendix

\section{Appendix / supplemental material}
\FloatBarrier

\subsection{Data and Code Availability}
We provide the source code and a dataset used in this study on Zenodo under \cite{echoPTZenodo}.

\FloatBarrier

\subsection{Extended Prediction Example}
\begin{figure}[H]
    \centering
    \includegraphics[width=\linewidth]{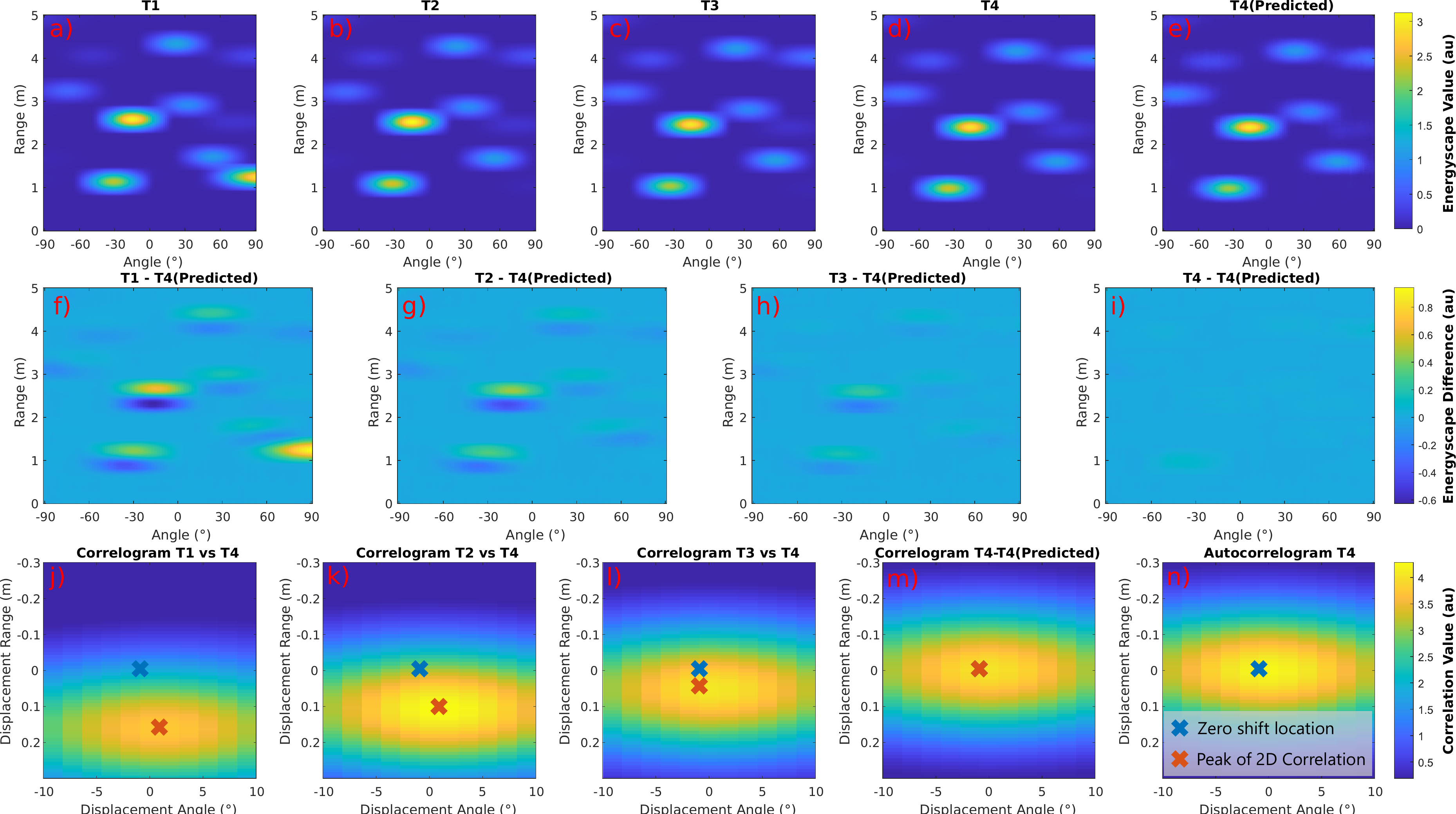}
    \caption{Detailed overview of some EchoPT predictions. Given a sequence of sonar images T1 to T4 (panels a-d) with a robot doing a linear motion in a corridor, the EchoPT model predicts T4(Predicted) in panel e. Panels f-i show the difference between T4(Predicted) with T1 to T4. These plots show that the model can capture the motion model of the sensor modality, as the errors between T4 and T4(Predicted) are near zero. The differences with the older images clearly show that the robot has learned to incorporate the sensor flow data. Panels j-n show the 2D correlograms between the prediction and the input data.} 
    \label{fig:prediction_corridor}
\end{figure} 

\FloatBarrier

\subsection{Detailed Auto-regressive Prediction Results}
\begin{figure}[H]
    \centering
    \begin{minipage}{\linewidth}
        \centering
        \includegraphics[width=\linewidth]{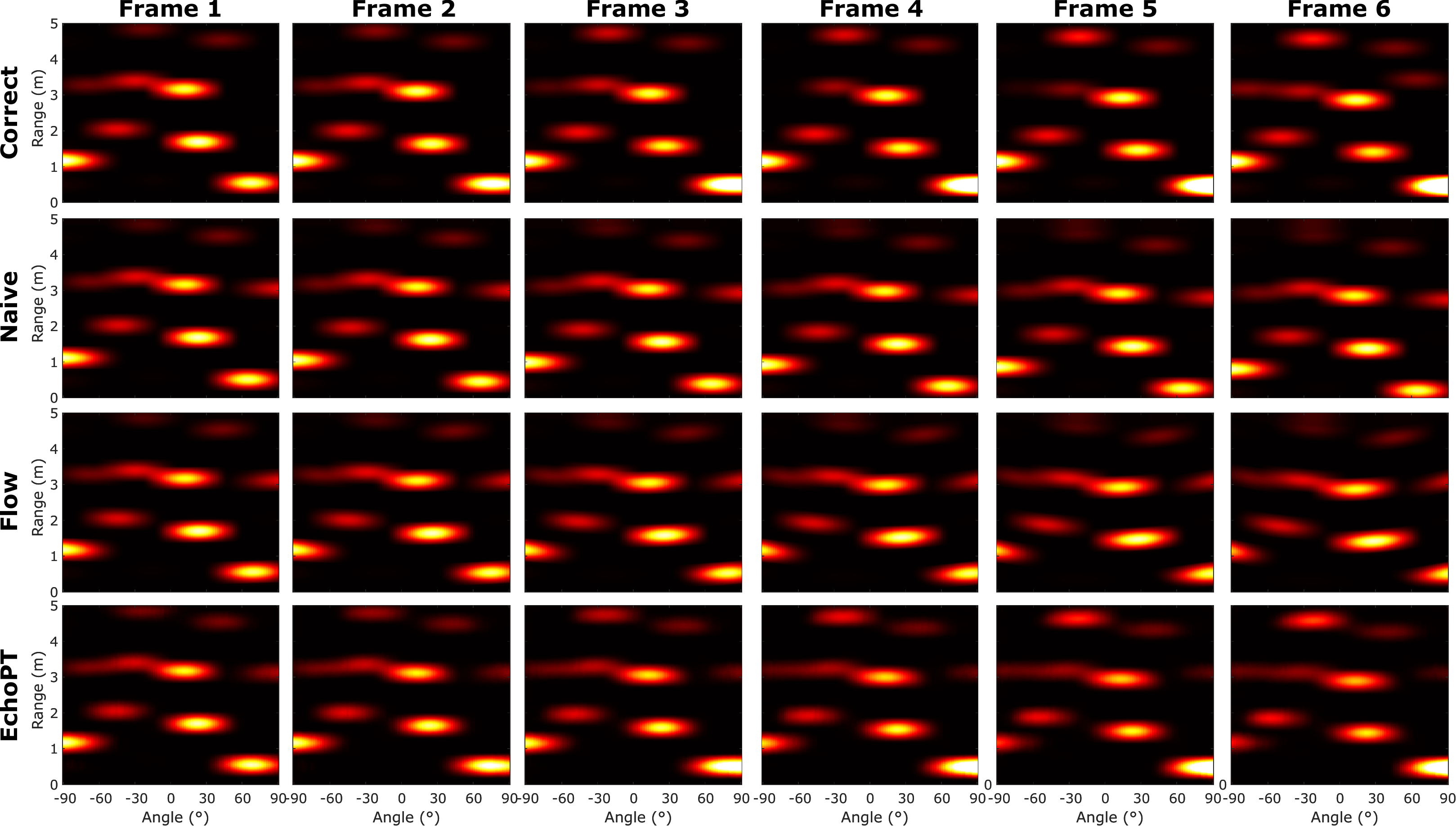}
        \caption{Prediction of sonar images using an auto-regressive prediction model, for the three prediction systems used in this paper (naive, acoustic flow and EchoPT). As the robot motions are relatively small, the difference between the images is not clearly visible. In figure \ref{fig:ar_diffs_appendix}, we show the differences between the subsequent images, as this illustrates much more clearly what the advantage of the EchoPT model is over the other techniques.}
        \label{fig:ar_methods_appendix}
    \end{minipage}
    
    \vspace{1cm} 
    
    \begin{minipage}{\linewidth}
        \centering
        \includegraphics[width=\linewidth]{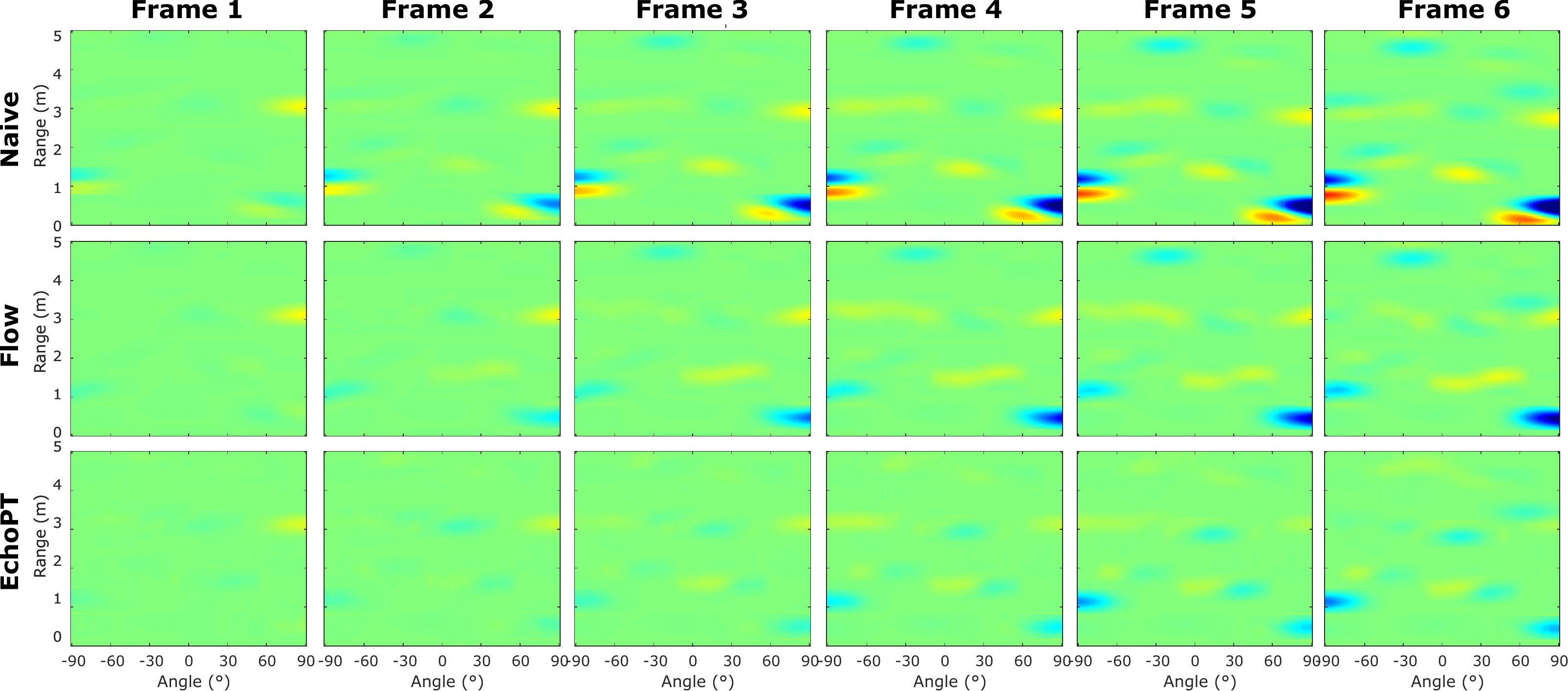}
        \caption{Prediction errors using an auto-regressive prediction model for the three prediction systems described. The deeper the prediction horizon, the larger the errors in the data predictions get (very noticeable in frame 6). The EchoPT model maintains the smallest prediction errors, indicating the capability of the model to predict over long time horizons. It should be noted that after frame 3, no measured data is used in EchoPT, but it purely relies on previous predictions to estimate the new data frame.}
        \label{fig:ar_diffs_appendix}
    \end{minipage}
\end{figure}

\FloatBarrier

\subsection{Details on the model architecture}
Here, we show the main model dimensions used in the EchoPT model. We encourage the reader to look at the source code found at \cite{echoPTZenodo} for more details.
\begin{table}[H]
\caption{Overview of the main model dimensions}
\label{tab:learnable_params}
\centering
\begin{tabular}{llc}
\toprule
Layer                 & Parameter                & Dimension             \\
\midrule
\textbf{embedding}    & Weights                  & 25×5×3×376             \\
                      & Bias                     & 376×1                  \\
\midrule
\textbf{posembed\_input} & Weights               & 376×2000               \\
\midrule
\textbf{TF Layers}    & QueryWeights             & 384×384                \\
(tf1 - tf8)           & KeyWeights               & 384×384                \\
                      & ValueWeights             & 384×384                \\
                      & OutputWeights            & 384×384                \\
                      & QueryBias                & 384×1                  \\
                      & KeyBias                  & 384×1                  \\
                      & ValueBias                & 384×1                  \\
                      & OutputBias               & 384×1                  \\
                      & conv1dProjection/Weights & 1×384×500              \\
                      & conv1dProjection/Bias    & 1×500                  \\
                      & conv1d/Weights           & 1×500×384              \\
                      & conv1d/Bias              & 1×384                  \\
\midrule
\textbf{LayerNorms}   & Offset                   & 1×1×3488               \\
                      & Scale                    & 1×1×3488               \\
\midrule
\textbf{batchnorm}    & Offset                   & 1×384                  \\
                      & Scale                    & 1×384                  \\
\midrule
\textbf{unembedLayer} & Wunembed                 & 384×125                \\
\midrule
\textbf{Conv Layers}  & Weights                  & 20×5×1×16, 20×5×16×32, 20×5×3×16, 20×5×65×16 \\
                      & Bias                     & 1×1×16, 1×1×32         \\
                      & Weights (transposed)     & 20×5×16×16, 20×5×4×16  \\
                      & Bias (transposed)        & 1×1×16, 1×1×4          \\
                      & Weights (conv2)          & 20×5×4                 \\
                      & Bias (conv2)             & 1×1                    \\
\midrule
\textbf{FC Layers}    & Weights (fc)             & 10×8                   \\
                      & Bias (fc)                & 10×1                   \\
                      & Weights (fc\_1)          & 10×10                  \\
                      & Bias (fc\_1)             & 10×1                   \\
                      & Weights (fc\_2)          & 200×10                 \\
                      & Bias (fc\_2)             & 200×1                  \\
\bottomrule
\end{tabular}

\end{table}

\FloatBarrier

\subsection{Optimizer Settings}
\begin{table}[H]
    \centering
    \label{tab:optim}.
    \caption{Optimizer Settings and Training Details}
    \begin{tabular}{@{}ll@{}}
        \toprule
        \textbf{Parameter} & \textbf{Value} \\
        \midrule
        Optimizer                    & Adam           \\
        GradientDecayFactor          & 0.9000         \\
        SquaredGradientDecayFactor   & 0.9990         \\
        Epsilon                      & 1.0000e-08     \\
        InitialLearnRate             & 5.0000e-05     \\
        MaxEpochs                    & 1000           \\
        LearnRateSchedule            & 'none'         \\
        LearnRateDropFactor          & 0.9700         \\
        LearnRateDropPeriod          & 10             \\
        MiniBatchSize                & 64             \\
        Shuffle                      & 'every-epoch'  \\
        CheckpointFrequency          & 30             \\
        CheckpointFrequencyUnit      & 'epoch'        \\
        SequenceLength               & 'longest'      \\
        PreprocessingEnvironment     & 'parallel'     \\
        L2Regularization             & 1.0000e-04     \\
        GradientThresholdMethod      & 'l2norm'       \\
        GradientThreshold            & Inf            \\
        Verbose                      & 1              \\
        VerboseFrequency             & 50             \\
        ValidationData               & []             \\
        ValidationFrequency          & 1000           \\
        ValidationPatience           & Inf            \\
        ObjectiveMetricName          & 'loss'         \\
        CheckpointPath               & '.'            \\
        ExecutionEnvironment         & 'auto'         \\
        OutputFcn                    & []             \\
        Metrics                      & []             \\
        Plots                        & 'training-progress' \\
        SequencePaddingValue         & 0              \\
        SequencePaddingDirection     & 'right'        \\
        InputDataFormats             & {'SSCB', 'CB'} \\
        TargetDataFormats            & 'auto'         \\
        ResetInputNormalization      & 1              \\
        BatchNormalizationStatistics & 'auto'         \\
        OutputNetwork                & 'best-validation-loss' \\
        Acceleration                 & 'auto'         \\
        \midrule
        GPU                          & 1x NVIDIA RTX 4090 \\
        Memory                       & 24 GB          \\
        Duration                     & 36 hours       \\
        Training set                 & 63000 input/output pairs \\
        Validation set               & 1600 sets      \\
        Test set                     & complete run of 25 min (7500 sets) \\
        \bottomrule
    \end{tabular}
\end{table}

\subsection{Animated Auto-regressive Prediction}
\begin{figure}[H]
    \centering
     \animategraphics[width=\textwidth, loop, autoplay]{2}{Figures/Animation1/anim-}{1}{51}
    \caption{This animation shows the auto-regressive prediction of sonar data using the EchoPT model. For the animation to work, view it with a PDF viewer that supports animations (such as Adobe PDF reader).}
    \label{fig:my_label}
\end{figure}

\end{document}